\documentclass[letterpaper, 10 pt, conference]{ieeeconf}  % Comment this line out if you need a4paper

\IEEEoverridecommandlockouts                              % This command is only needed if 
\usepackage{amsmath}
\usepackage{amssymb}
\usepackage{graphicx}
\usepackage[colorlinks=true, allcolors=blue]{hyperref}
\usepackage{caption}
\usepackage{subcaption}
\usepackage{gensymb}
\usepackage{multirow}
\usepackage{mathtools}
\usepackage{makecell}
\let\labelindent\relax % to solve the error of enumitem
\usepackage{enumitem}

\title{\LARGE \bf
A Learning-Based Approach for Estimating Inertial Properties of Unknown Objects from Encoder Discrepancies
}

\author{Zizhou Lao$^{1}$, Yuanfeng Han$^{2}$, Yunshan Ma$^{3}$ and Gregory S. Chirikjian$^{1}$% <-this % stops a space
\thanks{$^{1}$Zizhou Lao and Gregory S. Chirikjian are with the Department of Mechanical Engineering, National University of Singapore, Singapore (e-mail: lao.zizhou@u.nus.edu; mpegre@nus.edu.sg).}%
\thanks{$^{2}$ Yuanfeng Han is with the Department of Mechanical Engineering, Johns
Hopkins University, Baltimore, MD, USA (e-mail: yhan33@jhu.edu).}%
\thanks{$^{3}$ Yunshan Ma is with the Sea-NExT Joint Lab, National University of Singapore, Singapore (e-mail: yunshan.ma@u.nus.edu).}%
}

\begin{document}

\maketitle
\thispagestyle{empty}
\pagestyle{empty}

\begin{abstract}

Many robots utilize commercial force/torque sensors to identify inertial properties of unknown objects. However, such sensors can be difficult to apply to small-sized robots due to their weight, size, and cost. In this paper, we propose a learning-based approach for estimating the mass and center of mass (COM) of unknown objects without using force/torque sensors at the end-effector or on the joints. In our method, a robot arm carries an unknown object as it moves through multiple discrete configurations. Measurements are collected when the robot reaches each discrete configuration and stops. A neural network is designed to estimate joint torques from encoder discrepancies. Given multiple samples, we derive the closed-form relation between joint torques and the object's inertial properties. Based on the derivation, the mass and COM of object are identified by weighted least squares. In order to improve the accuracy of inferred inertial properties, an attention model is designed to generate weights of joints, which indicate the relative importance for each joint. Our framework requires only encoder measurements without using any force/torque sensors, but still maintains accurate estimation capability. The proposed approach has been demonstrated on a 4 degree of freedom (DOF) robot arm.

\end{abstract}

\section{Introduction}

In order to manipulate previously unseen objects, it is crucial for robots to infer physical properties such as shape, weight, material, and so forth \cite{billard2019trends, suomalainen2022survey, nguyen2019review, cui2021toward}. In this article, we develop a method for estimating mass and center of mass (COM) of a prior unknown objects being carried by robots.

Existing works in the field of robot manipulation have made great progress in estimating objects' inertial properties \cite{atkeson1985rigid, yu1999estimation, han2020can}. Another relevant topic is force estimation. Many attempts have been made to estimate interaction force of robots \cite{smith2005application, yilmaz2020neural, mattioli2016interaction}. However, there are still limitations of previous works. Firstly, commercial force/torque sensors are heavy and expensive, which are not commonly equipped on small-sized robots. In these scenarios, force/torque sensor based methods are not applicable. Secondly, many methods treat all the joints equally and then identify objects' properties using the measurements of joints. However, the distinctive information may be less concentrated or focused. In order to dynamically select important information, it is necessary to develop a mechanism that can evaluate the relative importance of each joint.

To address the above issues, we propose a learning-based framework to estimate mass and COM of unknown objects. Without using force/torque sensors, we only use encoders because they are light-weight, small and cheap, and are already built in to most robots \cite{wright2007design, hogg2002algorithms, gouaillier2009mechatronic}. In particular, we find that the discrepancy between the commanded joint angle and that observed by the encoder is useful in assessing load. To improve the estimated inertial properties, we design an attention model to evaluate the relative importance of each joint. Attention mechanism is a type of learning technique that adaptively generate weights for input information. The weight assigned to each piece of information is generally high if it is important, and low if it is unimportant. This process leads to a dynamic selection of information. Attention mechanisms have been widely used in the field of deep learning \cite{guo2022attention}, particularly in areas such as computer vision and national language processing, but less explored in the context of mechanics and robotics. Different from the previous attention mechanisms, we design a novel attention model suitable for the scenario of robots, where the weights assigned to joints for inertial properties inferring are adjusted dynamically. To the best of authors' knowledge, the use of attention mechanisms is introduced here for the first time to infer force/torque information indirectly from encoder discrepancies.

In the proposed framework, a neural network is designed to estimate joint torques in a robot arm. According to the estimated torque, the mass and COM of unknown objects held at the end-effector are solved analytically. An attention model is designed to generate weights for the joints, indicating the accuracy of torque estimation for each joint at a specific state. The main contributions of the proposed framework are as follows:
\begin{itemize}[leftmargin=*]
  \item A neural network is trained to estimate joint torques accurately with only the measurements from encoders, which saves the trouble of using force/torque sensors.
  \item For a robot carrying an object at steady state, the closed-form relationship between joint toque and the object's inertial properties is derived. Based on the derivation, mass and COM can be solved analytically by weighted least squares.
  \item An attention model is designed to generate weights assigned to joints dynamically, which helps to improve the accuracy of estimated inertial properties.
\end{itemize}

\begin{figure*}[t]
% \vspace{-1em}
\centering
\includegraphics[width=\textwidth]{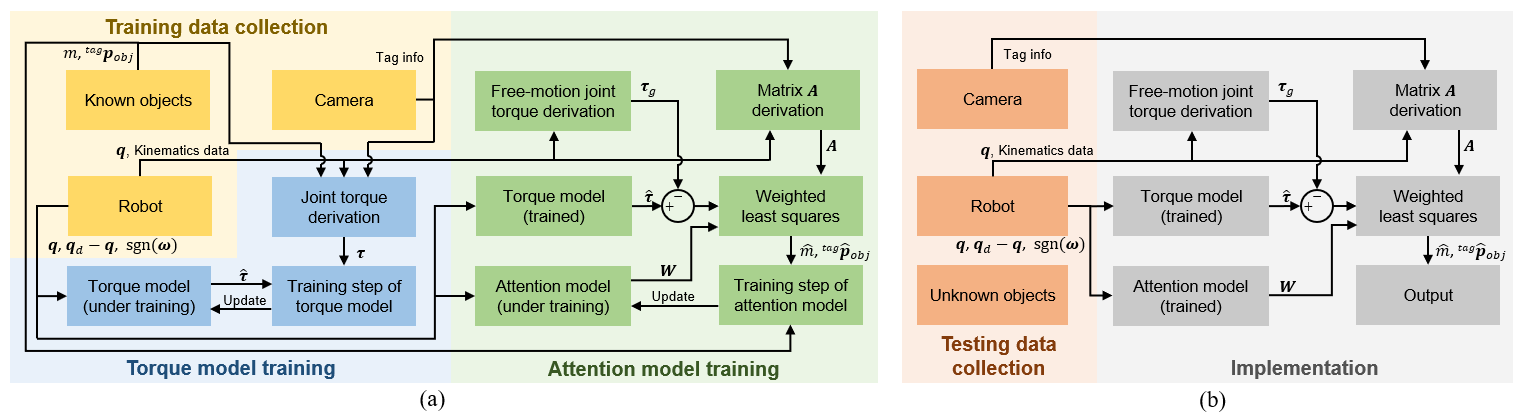}
\caption{Block diagram of the proposed approach. (a) Training process. The training data are collected with known objects. The torque model and attention model are trained sequentially. (b) Testing process. The testing data are collected with unknown objects. The torque model estimates the joint torque, and the attention model generates a weight matrix. The mass and COM are solved by weighted least squares.}
\label{fig:overview}
\vspace{-1em}
\end{figure*}

\section{Related Works}

We discuss existing works on the identification of inertial properties, force/torque estimation, and attention mechanisms.

\textbf{Identification of inertial properties.} In the field of robot manipulation, many methods are proposed to estimate inertial properties of unknown objects. For example, Atkeson et al. \cite{atkeson1985rigid} propose a method for estimating inertial parameters of a rigid body load from the measurements of a wrist force/torque sensor and arm kinematics. In \cite{yu1999estimation}, the mass and COM of an object is estimated by tipping and leaning operations. Based on a force sensing plate attached to the feet of humanoid robots, another approach for estimating physical properties of unknown boxes is proposed in \cite{han2020can}. The above methods require measurements of force/torque sensors, which are not commonly equipped on small-sized robots.

\textbf{Force/torque estimation.} Another relevant topic is robot interaction force or joint torque estimation, which is applicable to scenarios where force/torque sensors are unavailable. For example, Smith and Hashtrudi-Zaad \cite{smith2005application} and Yilmaz \cite{yilmaz2020neural} propose approaches for robot external force estimation. In these works, joint torque of free motion is estimated by neural networks, but motor torque sensors are still needed. In \cite{mattioli2016interaction}, joints' steady-state position error is utilized to reconstruct interaction force analytically for humanoid robots. There are also many works using neural networks to infer interaction forces from visual data \cite{kim2012image, hwang2017inferring} or video \cite{kim2019efficient}. However, most previous works lack the analysis of frictional torque and ignore the differences between joints.

\textbf{Attention mechanisms.} We design an attention model in the proposed framework to improve the performance. As one of the most important concepts in the fields of deep neural networks, attention mechanisms are widely used in various applications \cite{niu2021review}, such as natural language processing \cite{bahdanau2015neural, NIPS2017_3f5ee243} and computer vision \cite{guo2022attention, dosovitskiy2021an}. In the past few years, attention mechanisms have also been introduced to problems about robots \cite{lin2022efficient, wang2022heterogeneous, li2021message}. However, these works mainly utilize attention mechanisms to solve graphical problems, rather than problems in mechanics and robotics. Essentially, attention mechanisms are good at focusing on the distinctive parts when processing large amounts of information. We observed that this feature is suitable for our scenario, where the errors of joint torque estimation for individual joints are constantly changing. Hence, an attention model is designed to evaluate the weights of joints dynamically during robots' motions.

\section{Methods}

We consider a serial robot manipulator carrying an object moves through multiple configurations. For each configuration, the measurements are collected when all the joints reaches steady state. It is assumed that the joints of the robot are controlled through PD controllers, which have been adopted by many robots as their joint control strategy \cite{mattioli2016interaction, heredia2000high, robinson1999series}. A learning-based framework is proposed to identify the mass and COM of the object from multiple samples at steady state. Fig. \ref{fig:overview}(a) illustrates the training process of the proposed framework. By collecting multiple steady-state samples in experiments with several known objects, we train the torque model and attention model sequentially. Our framework is tested with several unknown objects as shown in Fig. \ref{fig:overview}(b).  For each sample, the joint torque is estimated by the torque model, and a weight matrix is generated by the attention model. According to the outputs of networks, the mass and COM of the unknown objects are solved analytically by weighted least squares.

\subsection{Problem Definition}

\begin{figure}[t]
% \vspace{-1em}
\centering
\includegraphics[width=1.0\linewidth]{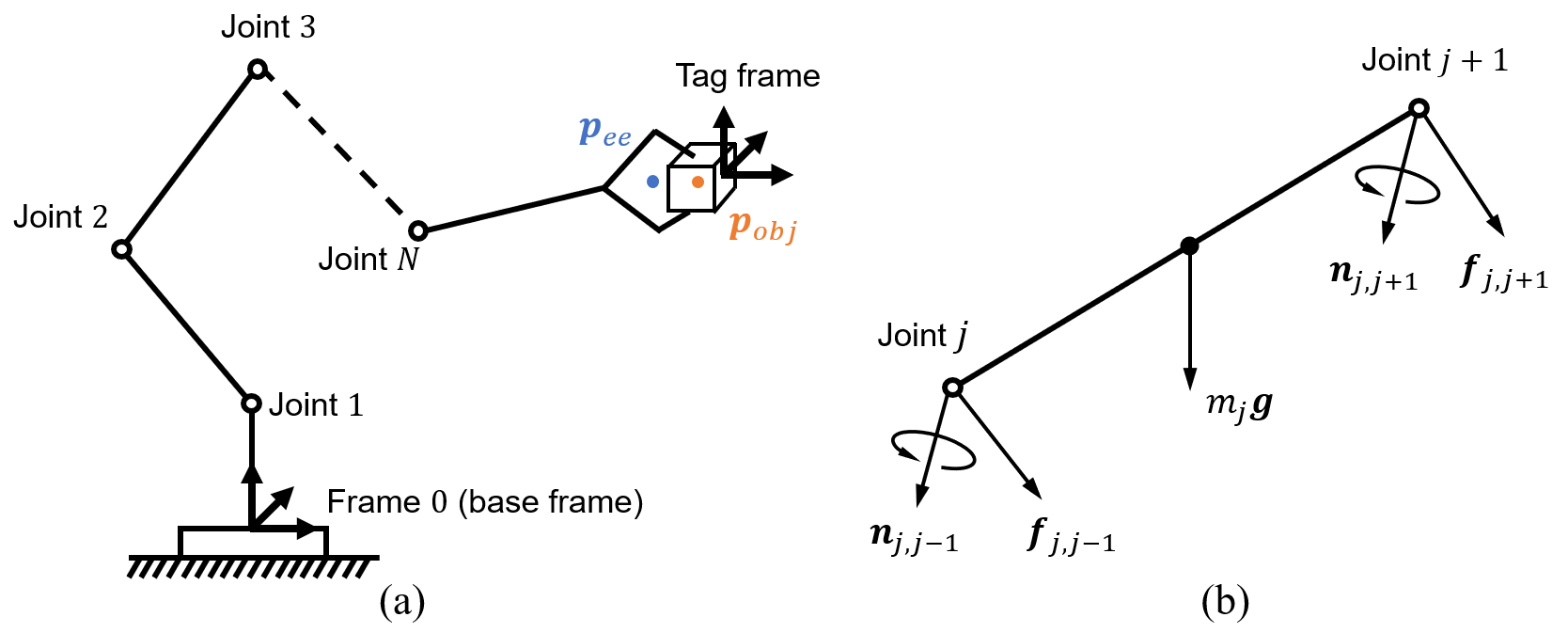}
\caption{(a) Schematic of an $N$-DOF robot carrying an object. (b) Free body diagram of the $j$-th link.}
\label{fig:robot}
\vspace{-1em}
\end{figure}

\begin{figure*}[t]

\centering
% \vspace{-1em} %调整图片与上文的垂直距离
\includegraphics[width=\textwidth]{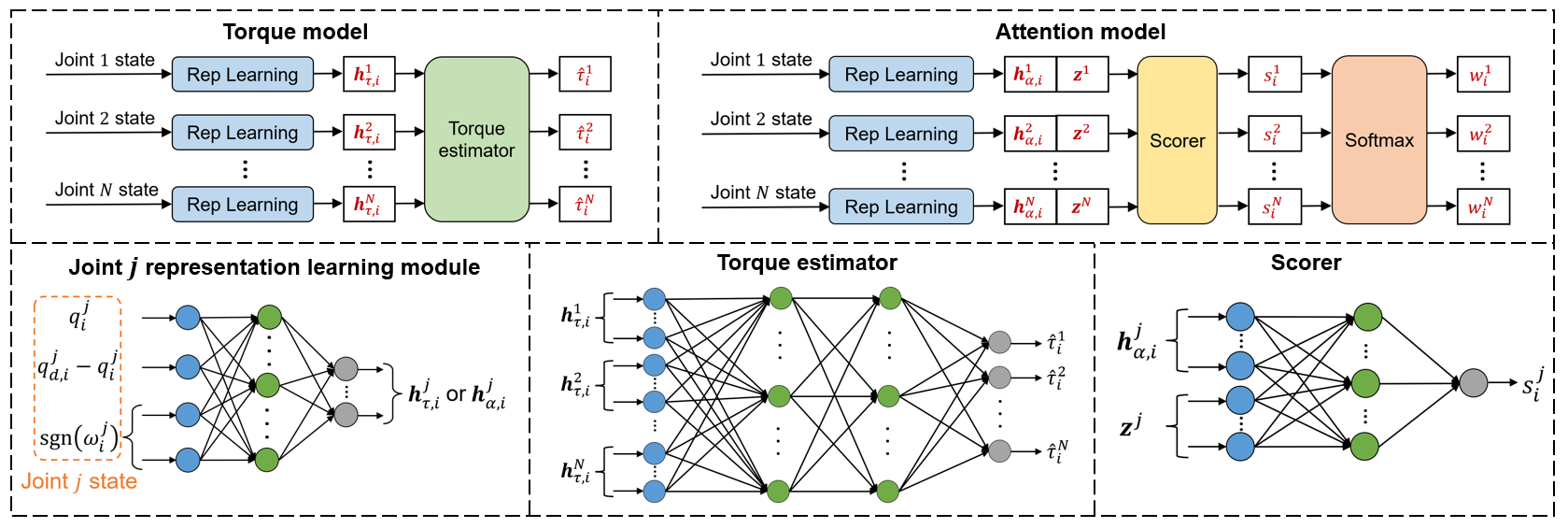}
\caption{The architectures of proposed neural networks. The first row illustrates the torque model and the attention model. The second row illustrates submodules in the above models.}
\label{fig:network}
\vspace{-1em} %调整图片与上文的垂直距离
\end{figure*}

Consider an $N$ degrees of freedom (DOF) robot carrying an unknown object as illustrated in Fig. \ref{fig:robot}(a). $M$ input samples are received to identify the mass and COM of the object. The ground truth value of mass is assumed to be $m$. A reference frame is assigned by sticking an April tag \cite{olson2011apriltag} to the object. The rigid body transformation from the tag frame to the robot base frame can be obtained by camera. Therefore, the ground truth of COM is represented by the 3-dimensional coordinates of COM in the tag frame $\prescript{tag}{}{\boldsymbol{p}}_{obj}$. Correspondingly, we denote the estimated inertial properties as $\hat{m}$ and $\prescript{tag}{}{\boldsymbol{\hat{p}}}_{obj}$. Regarding the $i$-th sample ($i=1, 2, \dots, M$) at steady state, the desired joint position and actual joint position are represented by $\boldsymbol{q}_{d,i}$ and $\boldsymbol{q}_i$, respectively. $\boldsymbol{\omega}_i$ refers to the joint angular velocity during the process of robot approaching steady state. We define the direction of rotation $\text{sgn}(\boldsymbol{\omega}_i)$ as the sign of $\boldsymbol{\omega}_i$. It should be noted that $\boldsymbol{q}_{d,i}$, $\boldsymbol{q}_i$ and $\text{sgn}(\boldsymbol{\omega}_i)$ are $N$-dimensional vectors, of which the $j$-th elements $q^j_{d,i}$, $q^j_i$ and $\text{sgn}(\omega^j_i)$ refer to the corresponding variables of joint $j$. We represent the desired joint position, actual joint position, and direction of rotation of all the $M$ samples as $\boldsymbol{q}_d$, $\boldsymbol{q}$, and $\text{sgn}(\boldsymbol{\omega})$, respectively.

\subsection{Neural Network for Estimating Joint Torque}

We design a neural network to estimate robot joint torque without force/torque sensors. The proposed torque model plays two roles: (i) reconstructing the motor torque from encoder discrepancies, and (ii) eliminating the effects of friction. Therefore, the output of torque model is the estimated joint toque corresponding to external force, including the gravitational force due to the weight of robot itself, as well as the interaction force at end-effector. 

We leverage the above insights in designing the torque model, which takes as input the joint position $\boldsymbol{q}$, joint position error $(\boldsymbol{q}_d  - \boldsymbol{q})$, and direction of rotation $\text{sgn}(\boldsymbol{\omega})$. For joints controlled through PD controllers, the motor torque is approximately proportional to the joint position error. So we take the joint position error as input for motor torque reconstruction. During the process of joint approaching steady state, the friction torque is along the opposite direction to rotation. Moreover, the magnitude of friction torque can be influenced by the joint position. Therefore, the direction of rotation and the joint angle are utilized to eliminate friction torque.

As shown in Fig. \ref{fig:network}, the torque model consists of $N$ joint representation learning modules and a torque estimator. A separate multi-layer perceptron (MLP) is utilized to embed the state of each joint. For the $i$-th sample and the $j$-th joint, the embedding is generated as:
\begin{equation}
    \boldsymbol{h}^j_{\tau, i} = \phi (q^j_i, q^j_{d, i} - q^j_i, \text{sgn}(\omega^j_i); \boldsymbol{\Theta}^j_{\text{rep}, \tau}),
\end{equation}
where $\phi(\cdot)$ refers to the representation learning module and $\boldsymbol{\Theta}^j_{\text{rep}, \tau}$ denotes the parameters of joint $j$'s representation learning in the torque model. Before embedding, $q^j_i$ and $q^j_{d, i}$ are normalized, and the direction of rotation $\text{sgn}(\omega^j_i)$ is converted to a 2-dimensional binary vector, i.e. 
$
\begin{bmatrix}
    1 & 0
\end{bmatrix}
$
for positive direction, and
$
\begin{bmatrix}
    0 & 1
\end{bmatrix}
$
for negative direction. Therefore, the input information of each joint state for representation learning is a 4-dimensional vector. After the representation learning process, the embeddings of all the joints are then concatenated, which models the interactions between joints. Another MLP is designed as the torque estimator, which takes as input the concatenated embedding and outputs estimated joint torque as:
\begin{equation}
    \boldsymbol{\hat{\tau}}_i =  \xi(\boldsymbol{h}^1_{\tau, i} \parallel \boldsymbol{h}^2_{\tau, i} \parallel \cdots \parallel \boldsymbol{h}^N_{\tau, i}; \boldsymbol{\Theta}_{\text{est}}),
\end{equation}
where $\xi(\cdot)$ refers to the torque estimator, $\boldsymbol{\Theta}_{\text{est}}$ denotes the parameters of the torque estimator, and $\parallel$ represents concatenation.

In the training process, the ground truth of joint torque can be calculated analytically. As the mass and COM of training object are known, we first calculate the force and moment applied on end-effector. Fig. \ref{fig:robot}(b) illustrates the free body diagram of the $j$-th link. $m_j \boldsymbol{g}$ indicates the gravitational force of link. $\boldsymbol{f}_{j,j-1}$ and $\boldsymbol{n}_{j,j-1}$ are the force and moment applied on the $j$-th link by the $(j-1)$-th link. And $\boldsymbol{f}_{j,j+1}$ and $\boldsymbol{n}_{j,j+1}$ are the force and moment on the $j$-th link by the $(j+1)$-th link. When the robot is stationary, the summation of force/moment exerted on the link is zero. Therefore, the forces and moments on all the joints from the end-effector to the base can be derived recursively by Newton's Laws \cite{beer1977vector}. Finally, the torque on each joint can be solved as the component of moment along the rotational axis. The ground truth of joint torque of all the samples is denoted as $\boldsymbol{\tau}$. Compared to the ground truth, we apply mean squared error (L2 loss) on the estimated joint torque to train the torque model as shown in Fig. \ref{fig:overview}(a).

Free-motion joint torque $\boldsymbol{\tau}_g$ is defined as the joint torque of robot at free motion, which is due to the weight of robot itself and irrelevant to the object at end-effector. We can calculate $\boldsymbol{\tau}_g$ recursively in a similar way as $\boldsymbol{\tau}$. The only difference is that the interaction force at end-effector is assumed to be zero. For each sample, the difference $(\boldsymbol{\tau}_i - \boldsymbol{\tau}_{g,i})$ is the portion of joint torque related to the interaction force at end-effector.

\subsection{Inferring Inertial Properties of Objects by Weighted Least Squares}

The closed-form relationship between joint torque and inertial properties are derived in this subsection. Based on the derivation, the mass and COM of object can be identified by weighted least squares, taking as input multiple steady-state samples. 

For the $i$-th sample, Jacobian matrix $\boldsymbol{J}_i \in \mathbb{R}^{6 \times  N}$ provides the relation between joint torque and interaction force at end-effector \cite{craig2005introduction} as
\begin{equation} \label{eq:J}
    \boldsymbol{\tau}_i - \boldsymbol{\tau}_{g, i} = \boldsymbol{J}_i^{\intercal} \boldsymbol{F}_i,
\end{equation}
where $(\boldsymbol{\tau}_i - \boldsymbol{\tau}_{g, i}) \in \mathbb{R}^N$ is the equivalent torque related to the endpoint force, and $\boldsymbol{F}_i \in \mathbb{R}^6$ denotes the wrench applied to the environment by the end-effector. In the case of robot carrying an object, the wrench can be written as
\begin{equation} \label{eq:wrench}
    \boldsymbol{F}_i = 
    \begin{bmatrix}
            -\boldsymbol{f}_i^{\intercal} & -\boldsymbol{n}_i^{\intercal}
    \end{bmatrix}^{\intercal},
\end{equation}
where $\boldsymbol{f}_i$ and $\boldsymbol{n}_i$ represent the force and moment exerted on end-effector by the object.

Given a single sample, the joint torque can be estimated through trained torque model, and the corresponding Jacobian can be calculated analytically. Therefore, it is possible to solve the wrench from Eq. \ref{eq:J} and use it for inferring object's inertial properties. Considering the scenarios when the DOF of robot is less than 6 or the robot is at singular configurations, as well as to improve the accuracy of inference, it would be better to take multiple samples as input. However, the wrench is not fixed for various robot postures. In order to process multiple samples, we need to directly build the relationship between joint torque and inertial properties.

When a robot carrying an object is at steady state, the force and moment exerted on end-effector are 
\begin{equation} \label{eq:f}
    \boldsymbol{f}_i = m \boldsymbol{g}, \\
\end{equation}
\begin{equation} \label{eq:mu}
    \boldsymbol{n}_i = (\prescript{0}{}{\boldsymbol{p}}_{obj,i} - \prescript{0}{}{\boldsymbol{p}}_{ee,i}) \times \boldsymbol{f}_i ,
\end{equation}
where the 3-dimensional $\boldsymbol{g}$ denotes gravitational acceleration, $\prescript{0}{}{\boldsymbol{p}}_{obj,i}$ and $\prescript{0}{}{\boldsymbol{p}}_{ee,i}$ denotes the coordinates of object's COM and end-effector in base frame, respectively. By sticking an April tag to the object as reference frame, the COM of object is represented by the coordinates in tag frame $\prescript{tag}{}{\boldsymbol{p}}_{obj}$. Then, the coordinates of object's COM in base frame can be obtained as
\begin{equation} \label{eq:p_obj}
    \prescript{0}{}{\boldsymbol{p}}_{obj,i} = \prescript{0}{tag}{\boldsymbol{R}}_i \cdot \prescript{tag}{}{\boldsymbol{p}}_{obj} + \prescript{0}{}{\boldsymbol{p}}_{tag,i},
\end{equation}
where $\prescript{0}{tag}{\boldsymbol{R}}_i$ is the rotation matrix from tag frame to base frame and $\prescript{0}{}{\boldsymbol{p}}_{tag,i}$ refers to the coordinates of tag in base frame. The above two terms can be obtained from camera. Substituting Eq. \ref{eq:f} and Eq. \ref{eq:p_obj} into Eq. \ref{eq:mu}, and converting cross product to matrix multiplication form, the moment can be written as
\begin{equation} \label{eq:mu_mul}
    \boldsymbol{n}_i = m [\boldsymbol{g}]_{\times}^{\intercal} \cdot (\prescript{0}{tag}{\boldsymbol{R}}_i \cdot \prescript{tag}{}{\boldsymbol{p}}_{obj} + \prescript{0}{}{\boldsymbol{p}}_{tag,i} - \prescript{0}{}{\boldsymbol{p}}_{ee,i}),
\end{equation}
where the skew-symmetric matrix $[\boldsymbol{g}]_{\times}$ is generated from the elements of $\boldsymbol{g}$ as
\begin{equation*}
    [\boldsymbol{g}]_{\times} =
    \begin{bmatrix}
        0 & -g_z & g_y \\
        g_z & 0 & -g_x \\
        -g_y & g_x & 0
    \end{bmatrix}.
\end{equation*}
Next, substituting Eq. \ref{eq:f} and Eq. \ref{eq:mu_mul} into Eq. \ref{eq:wrench}, the wrench at end-effector can be represented as
\begin{equation} \label{eq:wrench_x}
    \boldsymbol{F}_i = \boldsymbol{B}_i \boldsymbol{x},
\end{equation}
where the matrix $\boldsymbol{B}_i \in \mathbb{R}^{6 \times  4}$ is a function of joint position $\boldsymbol{q}_i$ and tag information as
\begin{equation} \label{eq:B}
    \boldsymbol{B}_i = 
    \begin{bmatrix}
        -\boldsymbol{g} & \boldsymbol{O} \\
        -[\boldsymbol{g}]_{\times}^{\intercal} \cdot ( \prescript{0}{}{\boldsymbol{p}}_{tag,i} - \prescript{0}{}{\boldsymbol{p}}_{ee,i}) & -[\boldsymbol{g}]_{\times}^{\intercal} \cdot \prescript{0}{tag}{\boldsymbol{R}}_i
    \end{bmatrix}
\end{equation}
and $\boldsymbol{x}$ is a 4-dimensional vector determined by the mass and COM of object as
\begin{equation} \label{eq:x}
    \boldsymbol{x} = 
    \begin{bmatrix}
        m & m \prescript{tag}{}{\boldsymbol{p}}_{obj}^{\intercal}
    \end{bmatrix}^{\intercal}.
\end{equation}
By substituting Eq. \ref{eq:wrench_x} into Eq. \ref{eq:J}, we can obtain the following equation:
\begin{equation} \label{eq:tau}
    \boldsymbol{\tau}_i - \boldsymbol{\tau}_{g,i} = \boldsymbol{A}_i \boldsymbol{x},
\end{equation}
where the matrix $\boldsymbol{A}_i \in \mathbb{R}^{N \times  4}$ is obtained as
\begin{equation}
    \boldsymbol{A}_i =  \boldsymbol{J}_i^{\intercal} \boldsymbol{B}_i.
\end{equation}
So far, we have extended the Jacobian to build the relation between between joint torque and inertial properties of object for a single sample. For multiple samples, the relations in Eq. \ref{eq:tau} can be synthetically written as
\begin{equation} \label{eq:tau_mul}
    \boldsymbol{\tau} - \boldsymbol{\tau}_g = \boldsymbol{A} \boldsymbol{x},
\end{equation}
where vectors $\boldsymbol{\tau} \in \mathbb{R}^{MN}$ and $\boldsymbol{\tau_g} \in \mathbb{R}^{MN}$, and matrix $\boldsymbol{A} \in \mathbb{R}^{MN \times 4}$ are generated by stacking the corresponding variables of $M$ samples as
\begin{equation}
    \boldsymbol{\tau} = 
    \begin{bmatrix}
        \boldsymbol{\tau}_1 \\
        \boldsymbol{\tau}_2 \\
        \vdots  \\
        \boldsymbol{\tau}_M
    \end{bmatrix}
    \boldsymbol{\tau_g} = 
    \begin{bmatrix}
        \boldsymbol{\tau}_{g,1} \\
        \boldsymbol{\tau}_{g,2} \\
        \vdots  \\
        \boldsymbol{\tau}_{g,M}
    \end{bmatrix}
    \boldsymbol{A} = 
    \begin{bmatrix}
        \boldsymbol{A}_1 \\
        \boldsymbol{A}_2 \\
        \vdots  \\
        \boldsymbol{A}_M
    \end{bmatrix}.
\end{equation}

In our framework, the estimated joint torque $\hat{\boldsymbol{\tau}}$ of multiple samples at steady state can be obtained by trained torque model. The corresponding free-motion joint torque $\boldsymbol{\tau}_g$ and matrix $\boldsymbol{A}$ can be calculated analytically. Assuming that the amount of samples $M$ is large enough so that Eq. \ref{eq:tau_mul} is overconstrained, we can obtain an optimal approximation of vector $\boldsymbol{x}$ by weighted least squares \cite{weisberg2005applied}. The cost function is defined as 
\begin{equation} \label{eq:c}
    C = (\boldsymbol{Ax} - (\hat{\boldsymbol{\tau}} - \boldsymbol{\tau}_g))^{\intercal} \boldsymbol{W} (\boldsymbol{Ax} - (\hat{\boldsymbol{\tau}} - \boldsymbol{\tau}_g)),
\end{equation}
where $\boldsymbol{W}  \in \mathbb{R}^{MN \times MN}$ is a diagonal weight matrix. The optimal estimation of $\boldsymbol{x}$ minimizing the cost function is
\begin{equation} \label{eq:x_hat}
    \boldsymbol{\hat{x}} = (\boldsymbol{A}^{\intercal} \boldsymbol{W} \boldsymbol{A})^{-1} \boldsymbol{A}^{\intercal} \boldsymbol{W} (\boldsymbol{\tau}-\boldsymbol{\tau}_g).
\end{equation}
Finally, the estimated mass and COM of the object can be solved from Eq. \ref{eq:x}.

\subsection{Assigning Weights to Joints by Attention Model}

In the process of inertial properties estimation, the cost function in Eq. \ref{eq:c} can be written as 
\begin{equation}
    C = \sum_{i=1}^{M} \sum_{j=1}^{N} w^j_i (\boldsymbol{A}^j_i \boldsymbol{x} - (\hat{\tau}^j_i - \tau^j_{g, i}))^2,
\end{equation}
where $\boldsymbol{A}^j_i \in \mathbb{R}^{1 \times 4}$ is the $j$-th row of $\boldsymbol{A}_i$, $\hat{\tau}^j_i$ is the $j$-th element of $\boldsymbol{\hat{\tau}}_i$, and $\tau^j_{g, i}$ is the $j$-th element of $\boldsymbol{\tau}_{g, i}$. Regarding the $i$-th sample and the $j$-th joint, $(\boldsymbol{A}^j_i \boldsymbol{x} - (\hat{\tau}^j_i - \tau^j_{g, i}))$ refers to the error between estimated torque and the torque derived from inertial properties of object. It can be seen that $C$ is the weighted sum of square error, where the element of weight matrix $w^j_i$ is the corresponding weight. 

We can simply set the weight matrix $\boldsymbol{W}$ to an identity matrix, which means all the joints are treated equally. However, it is better to adjust the weights dynamically as the joint torque errors can vary greatly in magnitude, for different joints or for different samples. For example, since the torque of a joint close to the end-effector is usually smaller than the torque of a joint close to the base, it is reasonable to increase the weights of the joint close to the end-effector appropriately. Moreover, when two joints are parallel, which means they provide similar information about the inertial properties of object, we can appropriately reduce the weight of the joint with larger torque error, so that the joint with smaller torque error contributes more.

In the proposed framework, we design an attention model to generate the weights of joints dynamically. As shown in Fig. \ref{fig:network}, the attention model consists of representation learning modules, a scorer and a softmax layer. Taking as input the $i$-th sample, the model outputs an $N$-dimensional weight vector $\boldsymbol{w}_i$, corresponding to the $N$ joints. Firstly, the joint states are embedded. Similar to torque model, the state includes joint position, joint position error and direction of rotation. For each joint, a separate MLP is designed as the corresponding representation learning module. The representation modules in torque model and attention model have the same architecture but the parameters are not shared. The embedding of the $j$-th joint and $i$-th sample can be represented as 
\begin{equation}
    \boldsymbol{h}^j_{\alpha, i} = \phi (q^j_i, q^j_{d, i} - q^j_i, \text{sgn}(\omega^j_i); \boldsymbol{\Theta}^j_{\text{rep}, \alpha}),
\end{equation}
where $\boldsymbol{\Theta}^j_{\text{rep}, \alpha}$ refers to the parameters of representation learning module of joint $j$ in the attention model. The indices of joints are then appended to the latent representations. The index of each joint is represented by an $N$-dimensional binary vector. For example,
$
\begin{bmatrix}
    0 & 0 & 1 & 0
\end{bmatrix}
$  
refers to the $3$-th joint of a $4$-DOF robot. Next, another MLP is introduced as a scorer to generate scores for all the joints according to the embeddings as 
\begin{equation}
    s^j_i = \gamma (\boldsymbol{h}^j_{a,i} \parallel \boldsymbol{z}^j; \boldsymbol{\Theta}_{\text{scorer}}),
\end{equation}
where $\gamma(\cdot)$ is the scorer, $\boldsymbol{z}^j$ denotes the index of joint $j$, $\boldsymbol{\Theta}_{\text{scorer}}$ denotes the parameters of the scorer, and $s^j_i$ is the score of joint $j$. Finally, the scores are normalized by a softmax function as 
$
    w^j_i = \frac{e^{s^j_i}}{\sum_{j=1}^{N} e^{s^j_i}}
$,
where the output $w^j_i$ denotes the weight of joint $j$ in the $i$-th sample. 

To estimate vector $\boldsymbol{x}$ by Eq. \ref{eq:x_hat} from $M$ samples, the attention model generates $M$ weight vectors for the corresponding samples. And the diagonal weight matrix $\boldsymbol{W}$ is generated as
$
    \boldsymbol{W} = \text{diag}(\boldsymbol{w}_1, \boldsymbol{w}_2, \dots, \boldsymbol{w}_M)
$,
of which the diagonal is the concatenation of all the weight vectors.

The attention model is trained after the torque model training process. The joint torque is first estimated by the trained torque model as shown in Fig. \ref{fig:overview}(a). Then the estimated inertial properties of objects are obtained by weighted least squares. In order to alleviate the influence of mass error on COM estimation, we use the ground truth of mass to solve COM from Eq. \ref{eq:x} in training process. While, in the testing process, COM is solved based on the estimated mass. Regarding the loss function, we apply L2 loss on both the estimated mass and COM. The loss for attention model training $L_\text{attention}$ is a weighted sum of mass loss $L_{m}$ and COM loss $L_\text{com}$ as
$
    L_\text{attention} = w_{m} L_{m} + w_\text{com} L_\text{com}
$,
where the weights $w_{m}$ and $w_\text{com}$ are manually set. 

\section{Experiments}

\begin{figure}[t]
% \vspace{-1em}
\centering
\includegraphics[width=\linewidth]{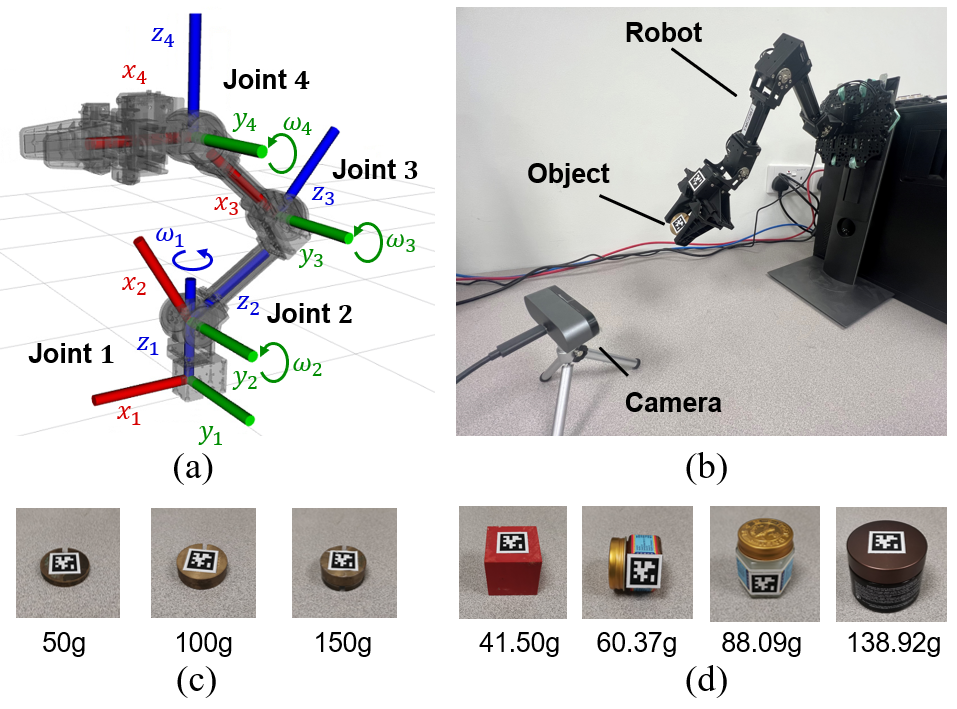}
\caption{(a) OpenMANIPULATOR-X robot manipulator. (b) Experimental setup. (c) Training objects. (d) Testing objects.}
\label{fig:experiment}
\vspace{-1em}
\end{figure}

\begin{table*}[t]
% \vspace{-1em}
\setlength{\abovecaptionskip}{-0.2cm} %调整图片标题与图距离
\caption{Error of estimated inertial properties. Sensor, PE, T-model and T-A model refer to the sensor based method, the position based method, proposed torque model without attention, and the proposed torque model with attention.}
\label{table:error_object}
\begin{center}
\resizebox{1.0\textwidth}{!}{
\begin{tabular}{c | c | c c c | c c c | c c c | c c c}
\hline \hline
& \multirow{2}{*}{Object} & \multicolumn{3}{c|}{\textbf{Sensor}} & \multicolumn{3}{c|}{\textbf{PE}} & \multicolumn{3}{c|}{\textbf{T-model}} & \multicolumn{3}{c}{\textbf{T-A model}} \\
& & MAE & NMAE & NRMSE & MAE & NMAE & NRMSE & MAE & NMAE & NRMSE & MAE & NMAE & NRMSE \\
\hline \hline
\multirow{5}{*}{\makecell {Mass error \\ (g / \% / \%)}} 
& Cube    & 4.54 & 10.94 & 13.55 & 12.00 & 28.92 & 32.12 & 2.49 & 6.00 & 7.59 & 2.40 & 5.79 & 7.27 \\ 
& Red     & 7.22 & 11.95 & 14.52 & 6.05 & 10.02 & 12.46 & 3.18 & 5.27 & 6.59 & 2.64 & 4.37 & 5.48 \\ 
& White   & 6.60 & 7.49 & 9.36 & 7.87 & 8.94 & 11.41 & 5.11 & 5.81 & 6.77 & 4.26 & 4.84 & 5.93 \\ 
& Black   & 15.69 & 11.30 & 12.84 & 8.92 & 6.42 & 8.05 & 10.18 & 7.33 & 7.66 & 5.57 & 4.01 & 4.74 \\  
\cline{2-14}
& Average & 8.51 & 10.42&	12.57	&8.71	&13.58	&16.01	&5.24	&6.10	&7.15&	3.72	&4.75	&5.86
\\  
\hline \hline
\multirow{5}{*}{\makecell{CoM error \\ (mm / \% / \%)}} 
& Cube    & 276.1 & 398.51 & 400.27 & 128.7 & 185.75 & 191.83 & 15.3 & 22.04 & 25.08 & 9.7 & 13.93 & 15.80 \\ 
& Red     & 193.2 & 290.24 & 291.13 & 103.2 & 155.00 & 156.94 & 12.6 & 18.87 & 20.98 & 9.3 & 13.92 & 15.47 \\ 
& White   & 143.5 & 189.97 & 190.74 & 76.7 & 101.55 & 103.10 & 7.8 & 10.38 & 11.98 & 6.5 & 8.62 & 9.98 \\ 
& Black   & 103.7 & 121.73 & 122.28 & 59.8 & 70.25 & 71.40 & 11.8 & 13.91 & 14.31 & 9.6 & 11.29 & 11.46 \\  
\cline{2-14}
& Average & 179.1	&250.11	&251.11&	92.1	&128.14	&130.82	&11.9	&16.30	&18.09	&8.8	&11.94&	13.18
\\  
\hline \hline
\end{tabular}}
\end{center}
\vspace{-1em}
\end{table*}

The proposed framework is verified on a 4-DOF robot OpenMANIPULATOR-X as shown in Fig. \ref{fig:experiment}(a). The joints from base to end-effector are joints 1, 2, 3 and 4. Fig. \ref{fig:experiment}(b) illustrates the experimental setup. The axis of joint 1 is in horizontal direction so that the torque is not constant to zero. The training and testing objects are shown in Fig. \ref{fig:experiment}(c) and (d). We attach April tags to the objects as reference frames.

% We use 3 objects to collect training data (Fig. \ref{fig:experiment}(c)), of which the weights are 50g, 100g and 150g. A cube (41.50g), a red bottle (60.37g), a white bottle (88.09g) and a black bottle (138.92g) are used as testing objects.

\textbf{Training Dataset.} The training data consists of planning samples and random samples. The planning samples are generated through the following steps: (i) The joint positions of samples are uniformly distributed with $10^{\circ}$ intervals on the joint space; (ii) For each joint position, all the $2^4=16$ possible directions of rotation are collected; (iii) The robot carries no object (at free motion), 50g, 100g and 150g to collect the above samples respectively, except for the samples that may collide. In addition, 9000 random samples are collected for each training object (including no object). In summary, we collect 82144 samples for training, including 46144 planning samples and 36000 random samples. Since each step of inertial properties inference requires multiple samples, we construct another training dataset for attention model, in which each data consists of 64 samples. The samples are randomly selected from the training samples.

\textbf{Evaluation metrics.} In order to assess the performance of the proposed approach for estimating mass, COM and joint torque, we use the mean absolute error (MAE), normalized mean absolute error (NMAE), and normalized root mean square error (NRMSE) defined as
\begin{equation}
    MAE = \frac{1}{n} \sum_{k=1}^{n} \left| \hat{y}_k - y_k \right|,
\end{equation}
\begin{equation}
    NMAE = \frac{\frac{1}{n} \sum_{k=1}^{n} \left| \hat{y_k} - y_k \right|}{y_\text{scale}}  \times 100\%,
\end{equation}
\begin{equation}
    NRMSE = \frac{\sqrt{\frac{1}{n} \sum_{k=1}^{n} \left| \hat{y}_k - y_k \right|^2}}{y_\text{scale}}  \times 100\%,
\end{equation}
where $\hat{y}_k$ and $y_k$ ($k = 1, 2, \dots, n$) are the estimated value and ground truth, and $y_{scale}$ is a scale value for normalization. The actual mass is used as the scale value when calculating the error of mass. With regard to the COM, the difference $\left| \hat{y}_k - y_k \right|$ in the above equations refers to the distance between the estimated COM and actual COM, and the scale value is the length of diagonal of the smallest cuboid that can enclose the object. To evaluate the estimated torque, the scale value is set to the maximum joint torque.

\textbf{Baselines.} We compare the proposed approach against the following two baselines:
\begin{itemize}[leftmargin=*]
    \item Current sensor based method. The measurements of joint torque are available, which is obtained from built-in electric current sensor. The raw measurement includes friction torque, resulting in significant error for inertial properties identification. By considering the friction torque and adding a constant, we optimize the built-in sensor for the $i$-th sample and $j$-th joint as
    $
        \hat{\tau}^j_{\text{sensor},i} = \tau^j_{\text{raw},i} - \text{sgn}(\omega^j_i) \tau^j_{f,1} + b^j_1
    $,
    where $\tau^j_{\text{raw},i}$ is the raw measurement, constant $\tau^j_{f,1}$ refers to the friction torque, and $b^j_1$ is a constant.
    \item Position error (PE) based method. This is an extension of \cite{mattioli2016interaction}. The original method assumes the joint torque is proportional to the joint position error. We optimize this method similarly as
    $
        \hat{\tau}^j_\text{PE} = P^j_\text{PE} (q^j_{d,i}  - q^j_i) - \text{sgn}(\omega^j_i) \tau^j_{f,2} + b^j_2
    $,
    where $P^j_\text{PE}$, $\tau^j_{f,2}$ and $b^j_2$ are constants. The constant coefficients in baselines are obtained by curve fitting using the same training data. For the two baselines, the inertial properties are solved using identity weight matrices.
\end{itemize}

\textbf{Implementation Details.} For each joint, min-max normalization is applied to joint position, so that the normalized joint position is within interval $[0, 1]$. Joint position error and joint torque are scaled down so that the magnitude is less than or equal to 1. In representation learning process, each joint state is embedded using a 2-layer MLP. The 4-dimensional input vectors are embedded as 12-dimensional vectors, and the hidden layer has 12 dimensions. The representation learning modules in the torque model and the attention model do not share parameters. The torque estimator is a 3-layer MLP. The dimensions of input layer is 48 and output layer is 4. Both the 2 hidden layers have 64 dimensions. The scorer is a 2-layer MLP with 32 neurons in hidden layer. Scalar scores are generated from 16-dimensional embeddings. The above modules are with ReLU non-linearity. All the parameters of models are randomly initialized. The torque model is trained with a batch size of 256 for 300 epochs with an initial learning rate of 0.0003. The attention model is trained with a batch size of 32 for 30 epochs with an initial learning rate of 0.0001. The weights in the loss of attention model are $w_{m} = 1$ and $w_\text{com} = 0.3$. We use Adam optimizer for training.

\subsection{Validation of Inertial Properties Estimation} \label{sec:prop}

We use four novel objects to validate the proposed framework. 1000 random samples for each object are collected. Similar to training dataset, 64 samples are randomly selected for a step of inertial properties inference. The testing process is shown in Fig. \ref{fig:overview}(b). To eliminate the potential bias of random results, the identification process is repeated 1000 times for each object. Table \ref{table:error_object} shows the results of mass and COM estimation. It can be seen that all the methods estimate the masses successfully. But the 2 baselines fail to identify the COM. The torque model without attention is able to estimate the COM. The results are further improved after adding the attention model.

\subsection{Evaluation of Torque Model and Attention Model}

\begin{table}[t]
% \vspace{-1em}
\setlength{\abovecaptionskip}{-0.2cm} %调整图片标题与图距离
\caption{Error of estimated joint torque. The units of MAE, NMAE and NRMSE are N$\cdot$mm, \% and \%.}
\label{table:error_torque_discrete}
\begin{center}
\resizebox{1.0\columnwidth}{!}{
\setlength{\tabcolsep}{0.6mm}{
\begin{tabular}{c | c c c | c c c | c c c} 
\hline \hline
& \multicolumn{3}{c|}{\textbf{Sensor}} & \multicolumn{3}{c|}{\textbf{PE}} & \multicolumn{3}{c}{\textbf{T-model}} \\
& MAE & NMAE & NRMSE & MAE & NMAE & NRMSE & MAE & NMAE & NRMSE \\
\hline \hline
Joint 1 & 35.12 & 8.41 & 11.36 & 28.66 & 6.87 & 8.99 & 11.35 & 2.72 & 3.67 \\ 
Joint 2 & 60.83 & 5.38 & 6.80 & 67.07 & 5.93 & 7.41 & 26.34 & 2.33 & 3.11 \\ 
Joint 3 & 53.94 & 10.11 & 12.01 & 45.58 & 8.54 & 10.30 & 9.85 & 1.85 & 2.57 \\ 
Joint 4 & 27.93 & 11.33 & 13.67 & 18.81 & 7.63 & 9.56 & 6.54 & 2.65 & 3.59 \\ 
\hline
Average & 44.45 & 8.81 & 10.96 & 40.03 & 7.24 & 9.07 & 13.52 & 2.39 & 3.23 \\  
\hline \hline
\end{tabular}}}
\end{center}
\vspace{-1em}
\end{table}

Using the random testing samples collected in Section \ref{sec:prop}, we tested the accuracy of estimated joint torque as shown in Table \ref{table:error_torque_discrete}. It can be seen that all the three methods are capable of estimating joint torque. Among them, the proposed torque model outperforms the 2 baselines.

\begin{table}[t]
% \vspace{-0.8cm} %调整图片与上文的垂直距离
\setlength{\abovecaptionskip}{-0.2cm} %调整图片标题与图距离
\caption{Mean weights assigned to joints.}
\label{table:weight}
\begin{center}
\begin{tabular}{c | c c c c} 
\hline \hline
& Joint 1 & Joint 2 & Joint 3 & Joint 4 \\ 
\hline \hline
Cube    & 0.0285 & 0.0033 & 0.0195 & 0.9488 \\ 
Red     & 0.0287 & 0.0031 & 0.0196 & 0.9486 \\ 
White   & 0.0320 & 0.0032 & 0.0218 & 0.9430 \\ 
Black   & 0.0353 & 0.0034 & 0.0260 & 0.9354 \\  
\hline \hline
\end{tabular}
\end{center}
\vspace{-1em}
\end{table}

In order to evaluate the performance of attention model, we calculate the mean weights of joints for all the testing objects, as shown in Table \ref{table:weight}. Obviously, the weights of joint 4 is much larger than the others. It meets our expectations as the MAE of joint 4 torque is significantly smaller than others. Larger weights prevent the contribution of joint 4 from being ignored. Moreover, as the axes of joints 2,  3 and 4 are parallel to each other, they actually provide similar information to solve the inertial properties. In particular, only the measurements of joint 1 could identify the location of COM along the axes of joints 2, 3 and 4. Therefore, although the error of joint 1 torque is considerable, the weight is still large enough to provide the distinctive information. We can also observe that the weights changes according to the masses of objects. For example, as the object becomes heavier, the weight of joint 4 decreases while the other weights increase. Regarding the torque model without attention, the results of cube is worse than others. Relatively speaking, the performance of torque model with attention is more even, as the estimation of different objects are of similar accuracy. In summary, the attention model improves the accuracy by adaptively assigning weights to joints, which adjusts the contributions of joints.

\subsection{Estimating Switching Forces along a Continuous Trajectory}

\begin{figure}[t]
% \vspace{-1em}
\centering
\includegraphics[width=\linewidth]{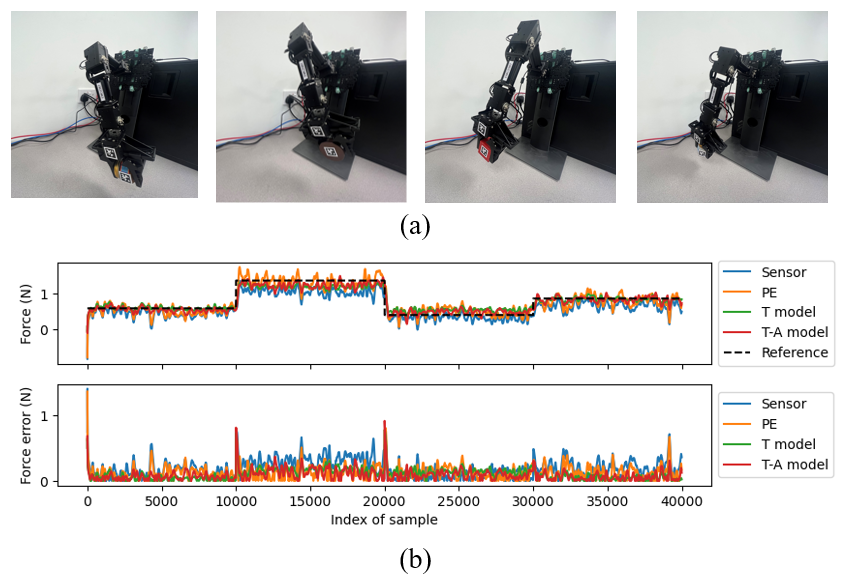}
\caption{(a) Snapshots of robot carrying objects. (b) Results of vertical force estimation in continuous experiments.}
\label{fig:continuous_force}
\vspace{-1em}
\end{figure}

\begin{figure*}[t]
% \vspace{-1em}
\centering
\includegraphics[width=\textwidth]{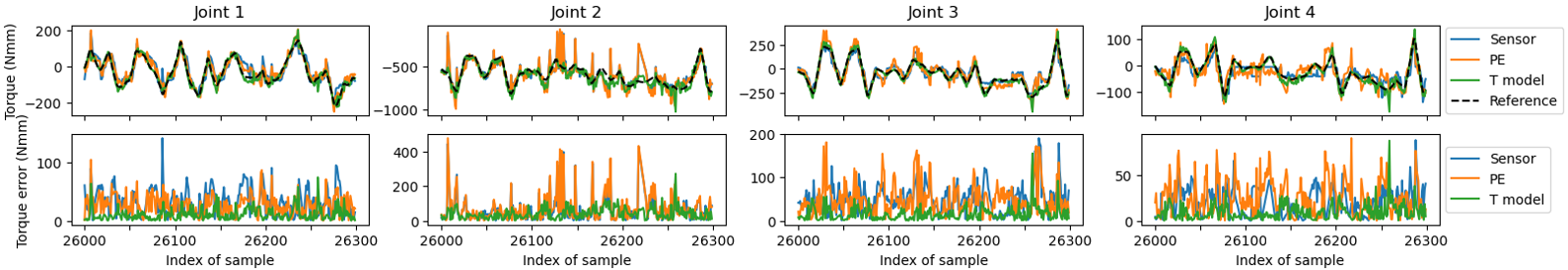}
\caption{Joint torque estimation in continuous experiments.}
\label{fig:continuous_torque}
\vspace{-1em}
\end{figure*}

Besides the above testing experiments, we did an experiment in the scenario of continuous trajectories with switching forces. The robot reaches a series of configurations. The measurements are collected at each configuration without stopping. During the motions, the robot carries switching objects, resulting in switching forces at end-effector (Fig. \ref{fig:continuous_force}(a)). Due to the proximity of the configurations, the samples provide similar information, which makes it challenging to accurately infer the inertial properties. To address this issue, 128 samples of consecutive configurations are used for each step of inertial properties identification. The vertical force is then computed from estimated mass. In addition, a 128-width mean filter is applied to the estimated force.

We notice that the proposed approach is able to estimate the switching forces as shown in Fig. \ref{fig:continuous_force}(b). It can be seen that the proposed method track the switching forces successfully. The torque model without attention outperforms the 2 baselines. And the attention model further reduces the force error. Parts of the joint torque results are plotted in Fig. \ref{fig:continuous_torque}. It can be seen that the proposed torque model is able to estimate consecutive postures along a trajectory, and outperforms the baselines.

\section{Conclusion}

A learning-based approach for estimating inertial properties of unknown objects is proposed in this paper. Without using force/torque sensors, we designed a torque model to reconstruct joint torque from encoder discrepancies. The closed-form relation between joint torque and inertial properties of objects are derived. Given multiple steady-state samples of robot carrying an object, the mass and COM of object can be solved analytically by weighted least squares. To adjust the weight matrix dynamically in the inference process, an attention model is designed to assign weights to joints. The proposed approach is verified in experiments on a 4-DOF robot manipulator. In conclusion, the proposed method achieves relatively accurate estimation of mass and COM without using force/torque sensors. 

\section*{Acknowledgment}

This work was supported by NUS Startup grants A-0009059-02-00 and A-0009059-03-00, CDE Board account E-465-00-0009-01, SMI Grant A-8000081-02-00, and National Research Foundation, Singapore, under its Medium Sized Centre Programme - Centre for Advanced Robotics Technology Innovation (CARTIN), sub award A-0009428-08-00.

\bibliographystyle{IEEEtran}
\bibliography{ref.bib}

\end{document}